\documentclass[a4paper]{jpconf}
\usepackage{graphicx}
\begin{document}
\title{Stacking Neural Network Models for Automatic Short Answer Scoring}

\author{Rian Adam Rajagede$^1$ and Rochana Prih Hastuti$^2$}
\address{$^1$ Department of Informatics, Universitas Islam Indonesia, Jalan Kaliurang km. 14,5, Sleman, Yogyakarta, Indonesia, 55584}
\address{$^2$ Department of Computer Science and Electronics, Universitas Gadjah Mada, Bulaksumur, Yogyakarta, Indonesia, 55281}
\ead{rian.adam@uii.ac.id}

\begin{abstract}
Automatic short answer scoring is one of the text classification problems to assess students' answers during exams automatically. Several challenges can arise in making an automatic short answer scoring system, one of which is the quantity and quality of the data. The data labeling process is not easy because it requires a human annotator who is an expert in their field. Further, the data imbalance process is also a challenge because the number of labels for correct answers is always much less than the wrong answers. In this paper, we propose the use of a stacking model based on neural network and XGBoost for classification process with sentence embedding feature. We also propose to use data upsampling method to handle imbalance classes and hyperparameters optimization algorithm to find a robust model automatically. We use Ukara 1.0 Challenge dataset and our best model obtained an F1-score of 0.821 exceeding the previous work at the same dataset.
\end{abstract}

\section{Introduction}
The short answer test is a type of exam in which participants are asked to answer questions with short answers which can consist of 1-2 sentences. Assessing student short answers on the exam is very challenging for the assessor. When a large number of students are assessed, for example on a national scale, assessors are required to remain consistent and objective in assessing hundreds or even thousands of student responses. The questions with short answer format also allow students to answer in their own style which can be varied for each student. Therefore, computer assistance in making automatic short answer scoring is deemed necessary to address these problems.

In 2019, NLP Research Group from Universitas Gadjah Mada, Indonesia, in collaboration with the Education Assessment Center, Ministry of Education and Culture of Indonesia, held the Ukara 1.0 Challenge\footnote{https://nlp.mipa.ugm.ac.id/ukara-1-0-challenge/}. In this challenge, participants from all over the nation were challenged to make an automatic short answer scoring system for Indonesian student exam. There are two short answer questions on this challenge with correct and incorrect labels. In this paper, we try to describe the improvement of our previous work on the Ukara 1.0 Challenge dataset.

There are several challenges that may arise in applying and making an automatic short answer scoring system. First, it is the number of models and hyperparameters that need to be searched.  In conventional machine learning, a model is trained to solve a specific problem, or in our case, a model is only trained to assess one short answer question. When the number of questions to be assessed increases, the time required for searching the model and tuning its hyperparameters will also increase. The second challenge is the imbalance class. In an exam, the questions given are intended to test students' abilities, therefore the level of difficulty in each question will cause the number of correct responses to be much less than the number of wrong responses. Another challenge in automatic short answer scoring is the small amount of labeled data. The data labeling process is not easy because it requires an expert to validate the student responses and of course this is time consuming and costly.

In several previous studies, making short answer scoring or essay scoring was done using deep learning approaches, for example, using long-short term memory, convolutional neural network, or a combination of both \cite{taghipour2016neural,riordan2017investigating,dong2017attention}. We take a different approach by using a simpler stacking model because of the small number of available data. In this paper, we used the sentence-level feature as done by \cite{dong2017attention}. Without word sequence features, the automatic scoring process could be viewed as a text classification problem. The use of stacking models for text classification has been done in \cite{abuhaiba2017combining,zhiwei2017email} and shows better performance than a single model classifier. We also propose to use an upsampling method, Synthetic Minority Over-sampling Technique (SMOTE)\cite{chawla2002smote}, to handle imbalance classes and hyperparameters optimization algorithm, Tree-structured Parzen Estimator (TPE)\cite{bergstra2011algorithms} to find a robust model that performs well on each type of question. In this paper, we use \textit{hyperparameters} term as model components which are preset and untrained (e.g. number of neural network layers). Meanwhile, the term \textit{parameters} refers to the model components that can be trained (e.g. neural network weights) \cite{deisenroth2020mathematics}.

\section{Dataset}
We use short answer scoring dataset from Ukara 1.0 Challenge Dataset published in 2019 by NLP Research Group Universitas Gadjah Mada. In this dataset, there are two different question sets, question set A and question set B. Each question set consists of a stimulus text, a question, and a collection of student responses with its label. Question A relates to natural disasters and question B is related to social problems. Examples of responses given by students along with its label are shown in table \ref{table1}.

\begin{table}
\caption{\label{table1}Examples of student response}
\begin{center}
\begin{tabular}{lll}
\br
Question&Response&Label\\
\mr
A& \begin{tabular}{@{}l@{}} mereka akan kehilangan lahan pertanian mereka \\ \textit{(they will lose their agricultural land)} \end{tabular} &Correct\\
A& \begin{tabular}{@{}l@{}} sulit menyesuaikan diri dengan keadan alam di tempat tersebut \\ \textit{(difficult to adjust to the natural conditions in that place)} \end{tabular} &Correct\\
A& \begin{tabular}{@{}l@{}} akan mendapat suasana baru \\ \textit{(will get a new atmosphere)} \end{tabular} &Incorrect\\
B& \begin{tabular}{@{}l@{}} karena akan bermanfaat bagi yg membutuhkan \\ \textit{(because it will be useful for those in need)} \end{tabular} &Correct\\
B& \begin{tabular}{@{}l@{}} karena dengan menyumbang para konsumen dapat membantu para pekerja \\ \textit{(because by donating consumers can help workers)} \end{tabular} &Correct\\
B& \begin{tabular}{@{}l@{}} karena mengurangi sampah lingkungan dan polusi \\ \textit{(because it reduces environmental waste and pollution)} \end{tabular} &Incorrect\\
\br
\end{tabular}
\end{center}
\end{table}

Each question set is also divided into three parts, namely the train set, dev set, and test set. This division refers to the time the competition takes place. Participants are only allowed to make models using the train set. Dev set is used to test at the validation stage where participants can only use it 20 times. While the test set is given at the final stage where participants are only allowed to test 5 times. we use the same scheme for the experiment process.

\subsection{Data Pre-processing}
There are several pre-processing techniques that is applied to the data before extracting the feature set. The first is case folding, which consists of converting all words to lowercase and case-folding the punctuation and unnormal words such as "\textit{yg}" to "\textit{yang}". This is to get better vector representation in the feature extraction process.

The second step is remove some frequent words. We do word counting for all vocab in the training data,  and  then  remove  some words based-on its frequency. We do not immediately eliminate most frequent words because it is worried that these words are important keywords in the scoring.

\subsection{Feature Extraction}
We propose to use two features to represent a student response. The first feature is sentence embedding extracted from Fasttext \cite{bojanowski2017enriching} using Indonesian fasttext pretrained model \cite{grave2018learning}. Fasttext pretrained model originally trained to get the subword-level embedding using CBOW architecture. In sentence embedding, the process of embedding occurs at the sentence level by calculating the average of word embedding value of each word as seen in (\ref{eq1}). Where $w_i$ is the i-th word embedding vector, N is the number of words in that sentence, and $\|w_i\|_2$ is L2 norm of vector $w_i$. We assume that each student's response is one sentence. When there are several sentences in one response, we process them as one sentence.

\begin{equation}
\label{eq1}
    s = \frac{1}{N}\sum_{i=1}^{N}\frac{w_i}{\|w_i\|_2}
\end{equation}

The second feature used is response’s word count. We assume that the length of the response matter to the scoring value. It is assumed that students with short responses convey less detailed information. For example. This feature was chosen following previous research by \cite{sultan2016fast}.The number of word in each response is normalized so that the value is between 0 and 1.

Using both features, each student's response is represented as a vector with a length of 301 values. The first 300 elements are result from sentence embedding and the last element is from the sentence length feature.

\subsection{Class Balancing}
It is natural in short answer questions that the number of correct responses is less than the number of incorrect responses. The  question  A  consists  of  191  correct  responses and 77 incorrect responses, while question B consists of 168 correct responses and 137 incorrect  responses.  It can be seen that the two questions have an imbalance number of classes. To balance this condition, we use Synthetic Minority Over-sampling Technique (SMOTE) \cite{chawla2002smote}.

\begin{figure}[h]
\includegraphics[width=23pc]{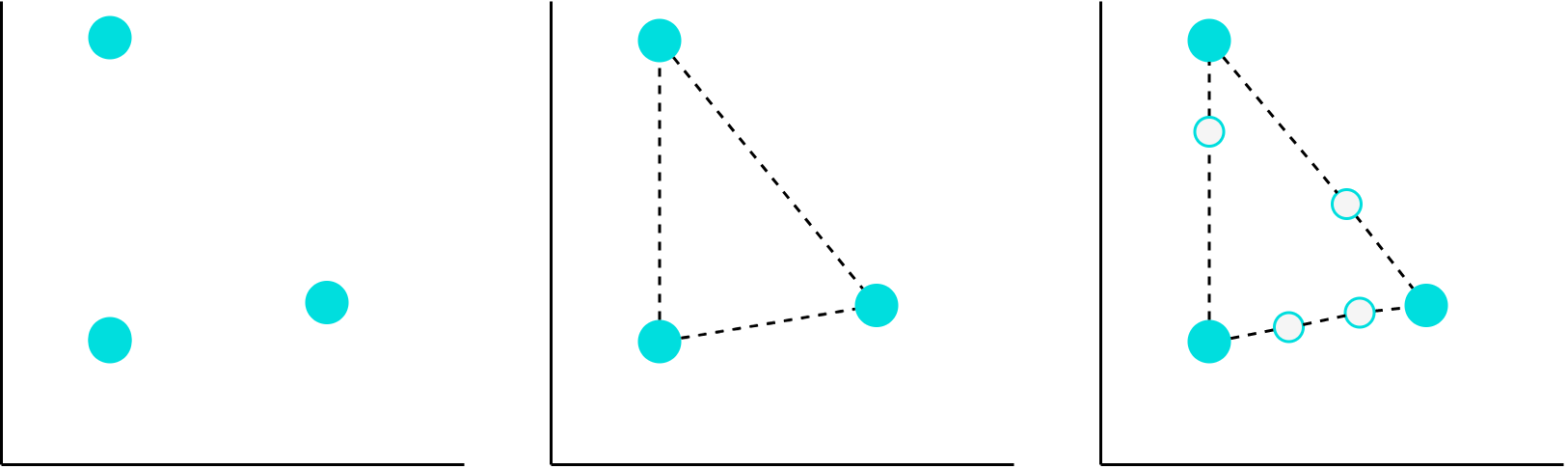}\hspace{2pc}%
\begin{minipage}[b]{12pc}\caption{\label{fig3}Illustration of SMOTE synthesizing new data.}
\end{minipage}
\end{figure}

SMOTE performs data upsampling based on the vector data in the feature space. As shown in figure \ref{fig3}, SMOTE synthesizes new data that is on the line between the example from the minority class. The blue dots are examples of data from the minority class, while the smaller dots are the new data. The synthetic data generated by SMOTE is data from correct responses with a total of 10\% of the total data. This proportion is less significant in numbers compared to the former proportion, while still fit the real case condition.

\section{Method}
For the classification process, we combined two types of classifier models, namely Extreme Gradient Boosting (XGBoost) \cite{chen2016xgboost} and neural network into a stacking model. We propose to use the stacking model using both classifiers as base classifier (level-0 model) and other neural networks as the final classifier (level-1 model). We avoid using deep learning model in this experiment because the amount of labeled data is small, which is only 268 responses to question A and 305 responses to question B. Besides, it is worried that a high computation resource needed for deep learning training process will become an obstacle during the implementation in the real world.

\subsection{Stacking model}
\begin{figure}[h]
\includegraphics[width=21pc]{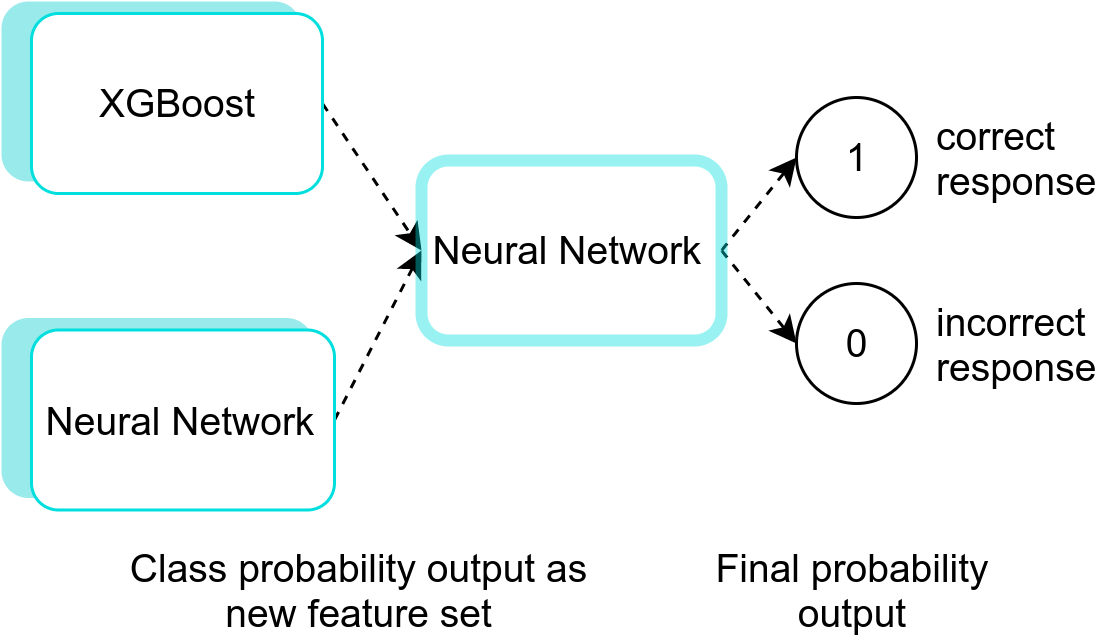}\hspace{2pc}%
\begin{minipage}[b]{14pc}\caption{\label{fig2}Stacking architecture using neural network and xgboost}
\end{minipage}
\end{figure}

Both  XGBoost  and  neural  networks  have  their own  advantages,  that is why we decide to combine these model in the stacking model \cite{wolpert1992stacked}. We use the stacking method, which generally use several classifier model to produce output that later to be used together for the next stacking layer. The output from the latest classifier model is the one that decide each label. In this work, we use output probability of each class from base model as an input for final model \cite{ting1999issues}. Therefore, XGBoost and neural network as base classifiers produce probability of each class for each student response and then this probability values is used as input in the next neural network model that is a final classifier. The architecture can be seen in figure \ref{fig2}.

Stacking models usually use a more complex model for the base model and a simpler model for the final model. We use deeper neural network on the base model (two hidden layers) and the shallow neural network in the final model (zero or one hidden layer). Both neural networks models optimized using Adam optimizer \cite{kingma}.

\subsection{Hyperparameters optimization}
In this work, hyperparameters searching process was conducted automatically using one of the automatic hyperparameters optimization algorithms, Tree-structured Parzen Estimator (TPE) \cite{bergstra2011algorithms}. TPE algorithm uses a bayesian approach $P(score | hyperparameters)$ to find the best hyperparameter. TPE will create a probability model based on the history of the hyperparameters that have been used and then choose the best hyperparameters value for the next tuning iteration. Several frameworks can be used for hyperparameters optimizations, for example, Hyperopt \cite{bergstra2013hyperopt} and Optuna \cite{akiba2019optuna}. In the experiment, we used Optuna as hyperparameters optimization framework with TPE algorithm.

\begin{figure}[h]
\includegraphics[width=23pc]{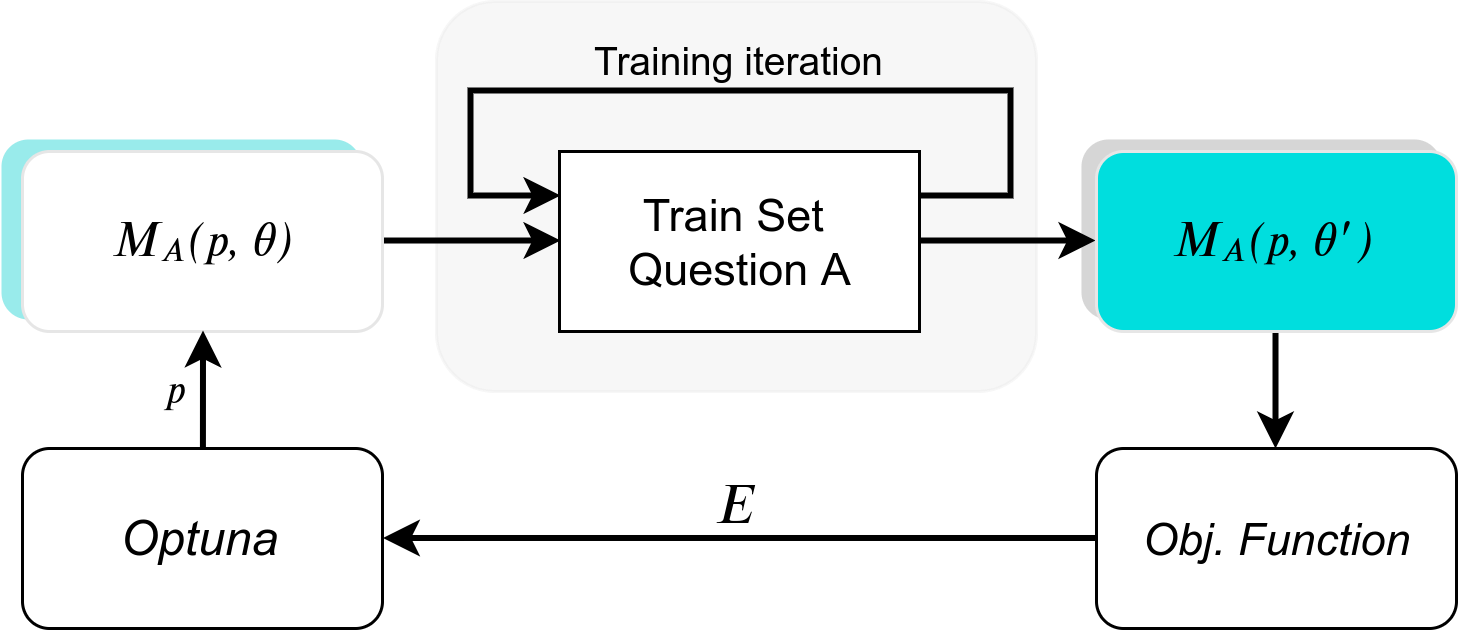}\hspace{2pc}%
\begin{minipage}[b]{12pc}\caption{\label{fig1}Training and hyperparameters tuning iteration using Optuna}
\end{minipage}
\end{figure}

Optuna will provide the set of hyperparameters $ p \in P $ where $ P $ is the defined search space of hyperparameters. The search space for each hyperparameter can be seen in the table \ref{table2}. The iteration process of training and tuning the model is shown in figure \ref{fig1}. Model $ M_A(p, \theta) $ is a model with the initial hypeparameters $ p $ and parameters $\theta$ for question A, while $ M_A(p, \theta) $ are the models that have been trained in the data for question A. The model is evaluated to calculate its F1-score, $E$, as objective function. Optuna will save the $(E, p)$ pair to update its probability model and try to maximize the F1-Score in the next iteration. We did 200 iterations of hyperparameters searching for each model.

\begin{table}
\caption{\label{table2}Search space of model hyperparameters}
\begin{center}
\begin{tabular}{ll}
\br
Hyperparameters & Search space \\
\mr
number of neuron in base neural net & $\{10k \mid k \in \{5, 75\} \}$ \\
value of neural net base learning rate & $\{1\times10^{-5}k \mid k \in \{2, 10000\} \}$ \\
base neural net iteration &  $\{10k \mid k \in \{1, 100\} \}$ \\
\mr
number of estimators in xgboost &  $\{2k \mid k \in \{25, 350\} \}$ \\
value of xgboost learning rate & $\{2\times10^{-5}k \mid k \in \{1, 5000\} \}$ \\
value of xgboost subsampling & $\{2\times10^{-2}k \mid k \in \{40, 50\} \}$ \\
\mr
number of layer in final neural net & $\{k \mid k \in \{0, 1\} \}$ \\
number of neuron in final neural net & $\{2k \mid k \in \{5, 100\} \}$ \\
value of neural net final learning rate & $\{1\times10^{-5}k \mid k \in \{1, 1000\} \}$ \\
final neural net iteration &  $\{10k \mid k \in \{1, 50\} \}$  \\
\br
\end{tabular}
\end{center}
\end{table}

The experimental scheme that we conducted follows the Ukara 1.0 Challenge scheme. The model is trained only using the train set. To avoid overfitting, we used k-fold cross validation by dividing the train data into five folds. Evaluation is done by calculating the average F1-score between five folds. Twenty best models tested using the dev set. Then the five models with the highest F1-score on the dev set were tested using the test set. The results written here are the best F1-score results from the test set. F1-score is calculated for each question using (\ref{eq2}).

\begin{equation}
\label{eq2}
    F1 = 2 \times \frac{(Precision \times Recall)}{(Precision + Recall)}
\end{equation}

\section{Results and discussion}
In the first experiment, we tried to compare the single model with the stacking model. We compare a single XGBoost model, a single neural network model, and a stacking model that combines both models. The hyperparameters of each model are from the same hyperparameters space as table \ref{table2}. The results of this experiment are shown in table \ref{table3}.

\begin{table}
\caption{\label{table3}Performance comparison of stacking and single model}
\begin{center}
\begin{tabular}{llll}
\br
Models & F1-Score A & F1-Score B & Combined F1-Score\\
\mr
Single model XGBoost & 0.88044 & 0.72695 & 0.80803 \\
Single model Neural Network & \textbf{0.88461} & 0.74343 & 0.81597 \\
Stacking model & 0.88192 & \textbf{0.75951}  & \textbf{0.81906} \\
\br
\end{tabular}
\end{center}
\end{table}

It can be seen in the table \ref{table3}, that the stacking model is better than the single models at the combined F1-Score. We use combined F1-score to evaluate the model against a combination of all test types at the same time. This combined evaluation is done by combining the predicted results of questions A and B and then calculating using (\ref{eq2}). This method was used in the Ukara 1.0 Challenge and the result may differ from averaging F1-score for all questions.

We also compare the effect of synthetic data on model performance. In table \ref{table4}, we can see that SMOTE improves the overall model performance. Especially in question A where the class ratio is more unbalanced than question B. In question B there was a decrease in the F1-score when using SMOTE.  This is probably due to the characteristics in question B which are more difficult than question A. It can also be seen from the large difference between the F1-Score of question A and question B. This decrease can also be due to the data ratio in question B which is sufficiently balanced and the use of SMOTE causes the model to be biased.

\begin{table}
\caption{\label{table4}Performance comparison of SMOTE usage}
\begin{center}
\begin{tabular}{llll}
\br
Models & F1-Score A & F1-Score B & Combined F1-Score\\
\mr
Stacking model & 0.88192 & \textbf{0.75951}  & 0.81906 \\
Stacking model + SMOTE & \textbf{0.88398} & 0.75895 & \textbf{0.82137} \\
\br
\end{tabular}
\end{center}
\end{table}

We compared the best model to the previously published methods at Ukara homepage. The results can be seen in table \ref{table5}. Unfortunately, on the Ukara homepage, the F1-Score results are displayed only up to two digits behind the comma. The two best models for this challenge use different approaches. The first author, using a deep learning method with bidirectional architecture LSTM with word embedding feature. Meanwhile, the second author used different methods for each question, random forest method for question A and logistic regression for question B. Our model outperformed both methods.

\begin{table}
\caption{\label{table5}Performance comparison of our method with previously published methods}
\begin{center}
\begin{tabular}{ll}
\br
Models  & Combined F1-Score \\
\mr
Bidirectional LSTM & 0.81 \\
Random Forest (A) and Logistic Regression (B) & 0.81 \\
Stacking model + SMOTE (ours) & \textbf{0.821} \\
\br
\end{tabular}
\end{center}
\end{table}

\section{Conclusion}
In this research, we have shown that the use of stacking model using neural network and xgboost makes it able to get a high score on the Ukara 1.0 challenge dataset. We also propose to use SMOTE and TPE to handle some problems in automatic short answer scoring and also improve the model performance. The model we produced received a combined F1 score of 0.821, better than previously published methods. From the results we obtained, it can be seen that there is still a big difference between F1 Score in question A and question B. This could be due to the different characteristics of the questions. In future research, it can be explored to use different features or methods for each question. Furthermore, it can also be explored the use of multi-task learning methods that train the model for all questions at the same time.

\section*{References}
\bibliography{iopart-num}

\end{document}